\def\year{2021}\relax
\documentclass[letterpaper]{article} 
\usepackage{aaai21}  
\usepackage{times}  
\usepackage{helvet} 
\usepackage{courier}  
\usepackage[hyphens]{url}  
\usepackage{graphicx} 
\usepackage{booktabs}
\usepackage{subfigure}
\usepackage{bbm}
\usepackage{amsmath}
\usepackage{multirow}
\usepackage{multicol}
\usepackage{natbib}
 \usepackage{amssymb}
\newcommand{\tabincell}[2]{\begin{tabular}{@{}#1@{}}#2\end{tabular}}
\usepackage{algpseudocode}  
\usepackage{algorithm}

\makeatletter
\def\@fnsymbol#1{\ensuremath{\ifcase#1\or \dagger\or \ddagger\or
   \mathsection\or \mathparagraph\or \|\or **\or \dagger\dagger
   \or \ddagger\ddagger \else\@ctrerr\fi}}
    \makeatother

\title{Tackling Instance-Dependent Label Noise via a Universal Probabilistic Model}
\author{
Qizhou Wang\textsuperscript{\rm 1,2}, 
Bo Han\textsuperscript{\rm 2,}\footnote{Corresponding authors. Emails: bhanml@comp.hkbu.edu.hk, chen.gong@njust.edu.cn. }, 
Tongliang Liu\textsuperscript{\rm 3}, 
Gang Niu\textsuperscript{\rm 4}, 
Jian Yang\textsuperscript{\rm 1},
Chen Gong\textsuperscript{\rm 1,5,}\footnotemark[1]\\}
\affiliations{
	\textsuperscript{\rm 1} Key Laboratory of Intelligent Perception and Systems for High-Dimensional Information of MoE, \\ School of Computer Science and Engineering, Nanjing University of Science and Technology \\
	\textsuperscript{\rm 2} Department of Computer Science, Hong Kong Baptist University\\
	\textsuperscript{\rm 3}  Trustworthy Machine Learning Lab, School of Computer Science, Faculty of Engineering, The University of Sydney\\
	\textsuperscript{\rm 4} RIKEN Center for Advanced Intelligence Project (AIP)\\
	\textsuperscript{\rm 5} Department of Computing, Hong Kong Polytechnic University \\
}
\begin{document}
	
	\maketitle
	
	\begin{abstract}
		The drastic increase of data quantity often brings the severe decrease of data quality, such as incorrect label annotations, which poses a great challenge for robustly training Deep Neural Networks (DNNs). Existing learning \mbox{methods} with label noise either employ ad-hoc heuristics or restrict to specific noise assumptions. However, more general situations, such as instance-dependent label noise, have not been fully explored, as scarce studies focus on their label corruption process. By categorizing instances into confusing and unconfusing instances, this paper proposes a simple yet universal probabilistic model, which explicitly relates noisy labels to their instances. The resultant model can be realized by DNNs, where the training procedure is accomplished by employing an alternating optimization algorithm. Experiments on datasets with both synthetic and real-world label noise verify that the proposed method yields significant improvements on robustness over state-of-the-art counterparts.
	\end{abstract}
	\section{Introduction}
	DNNs have gained much popularity in the literature of machine learning~\citep{he2017mask,jaderberg2015spatial}. However, one of the prominent factors for their success is the availability of large-scale training sample with clean {annotations}~\citep{deng2009imagenet}, where the acquisition process might be prohibitively expensive in practice. Hence,  various low-cost surrogate strategies are provided to automatically collect labels~\citep{li2017webvision,xiao2015learning}. These approaches make the acquisition of large-scale annotated data possible, but will also inevitably lead to noisy labels due to the imperfect results of search engines and crawling algorithms. Previous studies have suggested that noisy \mbox{labels} will hurt the test accuracy of DNNs~\citep{arpit2017closer,zhang2016understanding}. Thus, it would be desirable to develop robust methods for training DNNs with label noise. 
	
	In learning with noisy labels, existing robust learning \mbox{methods} consist of two general categories, according to whether the generation process for noisy labels is specified. The first category usually employs heuristic and ad-hoc rules without explicitly modeling the process of label corruption. For example, data cleansing techniques remove potential mislabeled instances before training~\citep{angelova2005pruning,sun2007identifying}; risk reweighting strategies demote the adverse impact of un-trustworthy \mbox{labels}~\citep{guo2018curriculumnet,han2018co,jiang2017mentornet,lee2018cleannet,wang2018iterative,han2020sigua}; label correction methods revise noisy labels based on model prediction~\citep{reed2014training,tanaka2018joint,yi2019probabilistic} or prior information~\citep{gao2017deep,li2017learning,veit2017learning}. These methods have gained extensive popularity in the literature. However, without  specifying the generation of label noise, these methods may lead to inferior or biased results~\citep{xia2019anchor,berthon2020confidence}.
	
	The above issue motivates us to explore the second category, which makes various assumptions on the process of label-noise generation. For example, the \textit{random classification noise} (RCN) model assumes that labels are randomly corrupted without any relationship with their real classes and instance features. To handle this simple situation, various noise-tolerant loss functions~\citep{ghosh2017robust,manwani2013noise,van2015learning} have been explored in the literature. Subsequently, the \textit{class-conditional noise} (CCN) model is proposed to relate noisy labels to true labels, \textit{i.e.}, some pairs of classes are more prone to be mislabeled. To depict such phenomenon, the noise transition matrix is introduced, which represents the probability of true labels flipping into noisy ones~\citep{patrini2017making,chen2020robustness}. This matrix plays a significant role in attacking CCN, and its elements can be either tuned by cross validation~\citep{natarajan2013learning} or estimated by various algorithms~\citep{liu2015classification,goldberger2016training,han2018masking,xia2019anchor,yao2020dual}.

	However, \cite{xiao2015learning} suggest that mis-annotation is highly related to instance features in reality. In other words, some instances may be confusing subjectively, and thus suffer from mislabeling. This brings us the \textit{instance-dependent noise} (IDN) model recently. To tackle this challenging problem, \cite{du2015modelling} consider the predetermined noise function that assumes mislabeled instances are close to the decision boundary; \cite{menon2016learning} propose an Isotron \mbox{algorithm} to decrease the difference between the clean and the corrupted distributions; \cite{cheng2017learning} cleanse data based on classification results on contaminated labels; \cite{berthon2020confidence} assume the probability that the assigned label is correct has been given; and \cite{xia2020parts} explore the instance-dependent transition matrix that is weighted combination of parts-dependent matrices. 
	Although these studies push the research field forward, their results are restricted to binary classification \mbox{problems} under various strong assumptions or rely on side information that may be unrealistic in practice, critically limiting their applications in real-world \mbox{situations}. Researchers also explore the underlying probabilistic process of label-noise generation, while the resultant probabilistic models rely on prior information \citep{xiao2015learning}, or count on specific model architecture \citep{goldberger2016training}. Therefore, how to  model the generation of IDN in a universal manner remains an important problem.
	
	\begin{figure}[t]
		\centering
		\subfigure[DOG]{
			\begin{minipage}[t]{0.33\linewidth}
				\centering
				\includegraphics[width=1in]{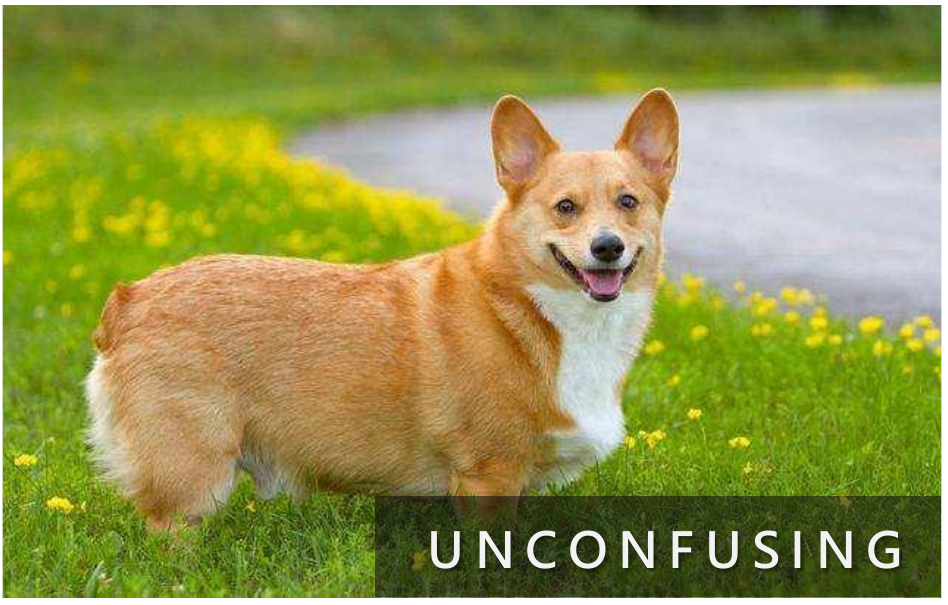}
				\label{figa: dog}
			\end{minipage}%
		}%
		\subfigure[DOG]{
			\begin{minipage}[t]{0.33\linewidth}
				\centering
				\includegraphics[width=1in]{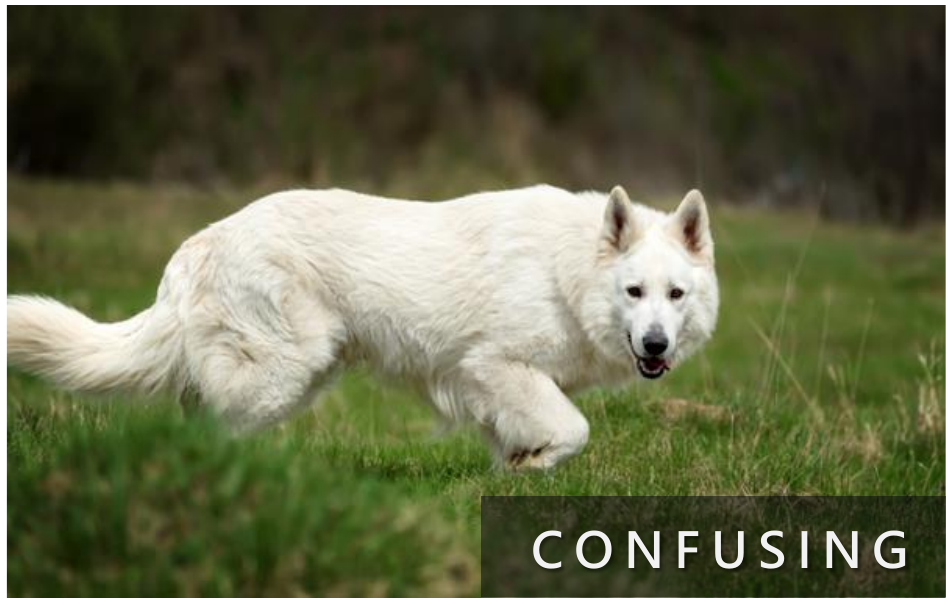}
				\label{figb: dog}
			\end{minipage}%
		}%
		\subfigure[WOLF]{
			\begin{minipage}[t]{0.33\linewidth}
				\centering
				\includegraphics[width=1in]{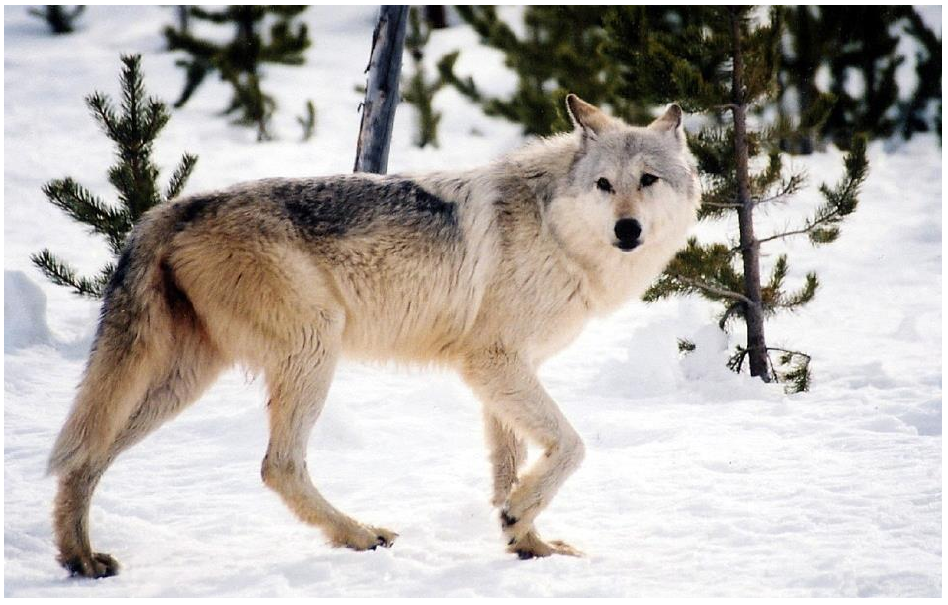}
				\label{figc: wolf}
			\end{minipage}
		}%
		\centering
		\caption{(a) and (b) should be both labeled as DOG. However, one may confuse about the label of (b), whose pattern resembles WOLF in (c).}
		\label{fig:dog and wolf}
	\end{figure}

\begin{figure}
	    \centering
	    \includegraphics[width=3.3in]{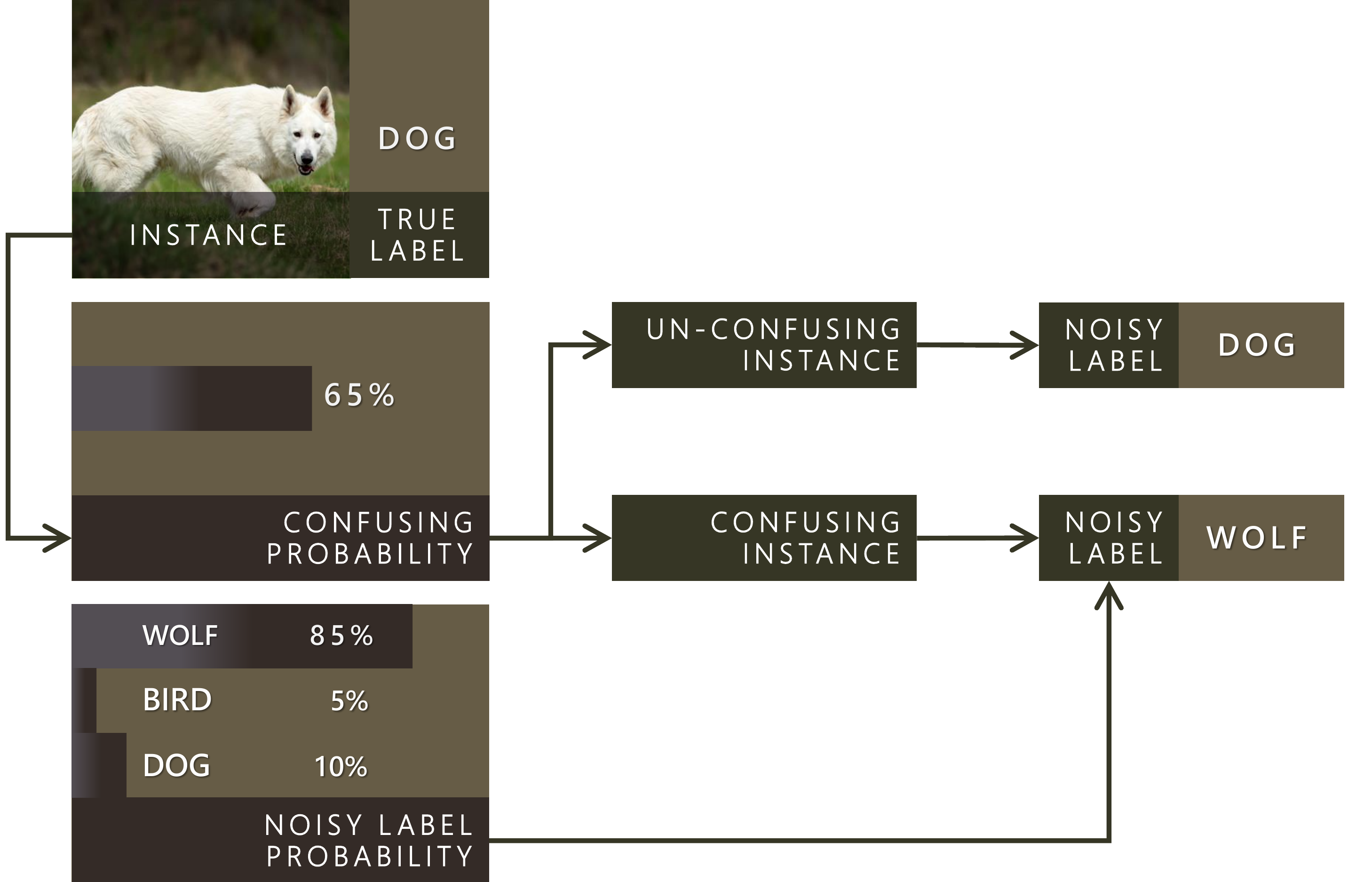}
	    \caption{An example of the generation of the noisy label. Due to the label confusing probability, the instance has 65\% to be a confusing instance. If it is confusing, the noisy label is generated by the noisy label probability itself, and is ``WOLF'' in this example. Conversely, if the instance is unconfusing, the noisy label will absolutely equal to the true label, \textit{i.e.}, ``DOG''.}
	    \label{fig:my_label}
	\end{figure}

	In this work, we focus on explicitly modeling the noisy label generation under the IDN assumption. Inspired by the \mbox{fact} that mislabeled instances often have particular patterns~\citep{xiao2015learning}, we assume that only those \textit{confusing} instances are suffering from mislabeling. For example, in Fig.~\ref{fig:dog and wolf}, (a) and (b)  should be both labeled as DOG. However, one may confuse about (b), whose pattern resembles WOLF in (c). Thus, (b) might be mislabeled, while (a) is easy to be correctly labeled.
	Accordingly, in distinguishing confusing and unconfusing instances, we propose a probabilistic IDN model to depict the label noise behavior (\textit{cf.}, Fig.~\ref{fig:my_label}). The model is realized by DNNs, and a novel alternating optimization algorithm is adopted to iteratively estimate true labels and update learnable parameters. Meanwhile, we investigate the learning behavior of our method that can find potentially confusing instances and correct their labels based on model prediction. We conduct experiments on \textit{CIFAR-100} and \textit{CIFAR-10} with synthetic label noise, and the empirical results demonstrate the state-of-the-art performance of our method. More importantly, we conduct real-world experiments on \textit{Clothing1M}~\citep{xiao2015learning}, where the noisy labels are critically instance-dependent. The empirical results indicate that our method also achieves the superior test accuracy over baselines in the real-world setting.
	
	\section{Tackling Instance-Dependent Label Noise}
	
	In this paper, the scalar is in lowercase letter (\textit{e.g.}, $a$), and the vector is in lowercase letter with boldface (\textit{e.g.}, $\mathbf{a}$) whose element can be accessed by superscript (\textit{e.g.}, $\mathbf{a}^j$). 
	
	Under the $c$-class problem setting, we define the label \mbox{space} $\mathcal{Y}=\left\{\mathbf{y} : \mathbf{y} \in\{0,1\}^{c}, \mathbf{y}^{\top} \mathbf{y}=1\right\}$ and the label distribution space $\mathcal{Q}=\left\{\mathbf{q} : \mathbf{q} \in [0,1]^c, \mathbf{1}^{\top} \mathbf{q}=1\right\}$. Specifically, for a label $\mathbf{y}\in\mathcal{Y}$ and the label distribution $\mathbf{q}\in\mathcal{Q}$, $\mathbf{y}^j=1$ indicates the instance is labeled to the $j$-th class, and $\mathbf{q}^j$ denotes the probability of $\mathbf{y}^j=1$. Note that each label $\mathbf{y}$ has only one non-zero value at the coordinate of the underlying class $j\in\left[c\right]$, where $\left[c\right]=\{1,\ldots,c\}$ is the set that contains all the categories. 
	
	\subsection{Label Noise Setting}
	
	
	In supervised classification, we consider the training sample $S=(\mathbf{x}_{1}, \ldots, \mathbf{x}_{N})$ drawn i.i.d. from some unknown distribution with the associated labels ${Y}=({\mathbf{y}}_{1}, \ldots, {\mathbf{y}}_{N})$, where $\mathbf{y}_i\in\mathcal{Y}$ is the one-hot label for the  instance $\mathbf{x}_i$. 
	However, the true labels $Y$ are inaccessible in the setting of label noise. Instead, we observe noisy labels $\widetilde{Y}=(\widetilde{\mathbf{y}}_{1}, \ldots, \widetilde{\mathbf{y}}_{N})$, where $\widetilde{\mathbf{y}}_i\in\mathcal{Y}$ might be different from the corresponding true label $\mathbf{y}_i$. 
	The task is to use the noisy-labeled sample $(S,\widetilde{Y})$ to \mbox{select} a softmax DNN classifier $\mathbf{h}_{\mathbf{w}}$ parameterized by $\mathbf{w}$, so that $\mathbf{h}_{\mathbf{w}}^j(\mathbf{x})$ properly estimates the true label probability $P(\mathbf{y}^j=1|\mathbf{x})$ for each new instance $\mathbf{x}$. 

	In order to tackle instance-dependent label noise, we begin by specifying our label noise setting. Concretely, instances are categorized into \textit{confusing} and \textit{unconfusing} instances, and we assume that only confusing instances suffer from mislabeling. Mathematically, an instance is named as a confusing instance if  $P(\widetilde{\mathbf{y}}_i=\mathbf{y}_i|\mathbf{y}_i,\mathbf{x}) < 1$, and as an unconfusing instance if  $P(\widetilde{\mathbf{y}}_i=\mathbf{y}_i|\mathbf{y}_i,\mathbf{x}) = 1$. To facilitate the derivation of our probabilistic model, we introduce a binary variable $s_i\in\{0,1\}$ for each instance $\mathbf{x}_i$, such that $s_i=1$ indicates the instance $\mathbf{x}_i$ is confusing, and $s_i=0$ otherwise. 
	
	As aforementioned, given an unconfusing instance $\mathbf{x}_i$, the noisy label $\widetilde{\mathbf{y}}_i$ {equals} to the true label $\mathbf{y}_i$ definitely, \textit{i.e.},
	\begin{equation}
	P(\widetilde{\mathbf{y}}_i|\mathbf{y}_i,\mathbf{x}_i, s_i=0)=\mathbb{1}\{\mathbf{y}_i=\widetilde{\mathbf{y}}_i\},
	\label{eq:noisy label c=0}
	\end{equation}
	where $\mathbb{1}\{\cdot\}$ is the indicator function that equals to 1 if the condition is true, and 0 otherwise. On the contrary, the noisy label $\widetilde{\mathbf{y}}_i$ might be incorrect if the instance is confusing. In this case, we assume that the noisy label is independent on the true label, namely,
	\begin{equation}
	P(\widetilde{\mathbf{y}}_i|\mathbf{y}_i,\mathbf{x}_i,s_i=1)=P(\widetilde{\mathbf{y}}_i|\mathbf{x}_i). \label{eq:noisy label c=1}
	\end{equation}
	It means that annotators fail to determine the true label, and thus a confusing instance cannot be properly classified according to annotators' knowledge. Fig.~\ref{fig:my_label} summarizes the key concepts of our IDN setting. Accordingly, the posterior of a noisy label $\widetilde{\mathbf{y}}_i$ can be summarized by:
	\begin{equation}
	\begin{aligned}
	P(\widetilde{\mathbf{y}}_i | {\mathbf{y}}_i, \mathbf{x}_i)
	\overset{1}{=} & \sum_{s\in\{0,1\}} P(\widetilde{\mathbf{y}}_i, s_i=s| {\mathbf{y}}_i, \mathbf{x}_i)\\
	\overset{2}{=} & \sum_{s\in\{0,1\}} P(\widetilde{\mathbf{y}}_i| {\mathbf{y}}_i, \mathbf{x}_i, s_i=s)P(s_i=s|\mathbf{x}_i)\\
	\overset{3}{=} & (1-\eta_i)\mathbb{1}\{\mathbf{y}_i=\widetilde{\mathbf{y}}_i\}+\eta_i \psi_i, \label{eq: main}
	\end{aligned}
	\end{equation}
	where we define $\psi_i=P(\widetilde{\mathbf{y}}_i|\mathbf{x}_i)$ and term $\eta_i=P(s=1|\mathbf{x}_i)$ as the \emph{confusing probability}. Note that, the $2^\text{nd}$ equation is based on the assumption $P(s_i=s|\mathbf{y}_i, \mathbf{x}_i)=P(s_i=s|\mathbf{x}_i)$, 
	and the $3^\text{rd}$ equation is derived by Eq.~\eqref{eq:noisy label c=0} and Eq.~\eqref{eq:noisy label c=1}. 
	
	\noindent \textbf{Remarks}. Different from previous works \citep{yan2014learning} that find incorrectly labeled instances, we categorize instances into confusing and unconfusing instances. Without proper information about true labels, finding confusing instances is much easier than finding mislabeled ones (\textit{cf.}, Fig.~\ref{fig:dog and wolf}). This merit can also simplify the resulting probabilistic model, since we do not require to take \mbox{different} true-label cases into account. Moreover, the independency assumption for confusing instances in Eq.~\eqref{eq:noisy label c=1} can also be extended to true label dependent cases~\citep{xiao2015learning}. However, our simple assumption leads to lucid and explicable learning method in Algorithm~\ref{algorithm}.
	
	\subsection{Objective Function}
	
	In addition to parameters $\mathbf{w}$ of the classifier $\mathbf{h}_\mathbf{w}$, we need to estimate $\psi_i$ and $\eta_i$ for each instance $\mathbf{x}_i$. The probability of the noisy label $\psi_i$ can easily be estimated by training a naive classifier on the original dataset with noisy labels~\citep{berthon2020confidence}. By contrast, each confusing probability $\eta_i$ has to be taken as a learnable parameter, and thus we aim to fit all trainable parameters $\Theta=\{\mathbf{w}, \eta_1, \ldots, \eta_N\}$ given only the noisy-labeled training sample $(S, \widetilde{Y})$. 
	One common approach adopted for selecting \mbox{parameters} $\Theta$ is based on the maximum likelihood principle, which aims to find the optimal parameters by maximizing the log-likelihood:
	\begin{equation}
	\begin{aligned}
	\ell(\Theta)&=\sum_{i\in\left[N\right]} \log \hat{P}(\widetilde{\mathbf{y}}_i | \mathbf{x}_i; \Theta)\\
	&=\sum_{i\in\left[N\right]} \log \sum_{j\in\left[c\right]} \hat{P}(\widetilde{\mathbf{y}}_i, \mathbf{y}^j_i=1 | \mathbf{x}_i; \Theta), \label{eq: likelihood function}
	\end{aligned}
	\end{equation}
	where $\hat{P}(\cdot)$ denotes the estimated probability. In the next section, we describe the optimization algorithm for Eq.~\eqref{eq: likelihood function}.
	
	Before moving to the next section, we provide a simple derivation that is useful in below:
	\begin{equation}
	\begin{aligned}
	\hat{P}(\widetilde{\mathbf{y}}_i, \mathbf{y}_i^j=1|\mathbf{x}_i;\Theta)
	{=} & \hat{P}(\widetilde{\mathbf{y}}_i|\mathbf{y}_i^j=1, \mathbf{x}_i;\eta_i)\mathbf{h}_{\mathbf{w}}^j(\mathbf{x}_i) \\
	{=} & \left[(1-\eta_i)\widetilde{\mathbf{y}}_i^j+\eta_i \psi_i\right]\mathbf{h}_{\mathbf{w}}^j(\mathbf{x}_i).
	\end{aligned} \label{eq: assist}
	\end{equation}
	The first equation is given by the definition of our target classifier, \textit{i.e.,} $\mathbf{h}_{\mathbf{w}}^j(\mathbf{x}_i)=\hat{P}(\mathbf{y}_i^j=1|\mathbf{x}_i;\mathbf{w})$; and the second equation is based on Eq.~\eqref{eq: main}, where $\widetilde{\mathbf{y}}_i^j$ is in place of $\mathbb{1}\{{\mathbf{y}}_i=\mathbf{\widetilde{y}}_i\}$ 
	since the vectors $\mathbf{y}_i,\widetilde{\mathbf{y}}_i$ each have only one non-zero element and we already have $\mathbf{y}_i^j=1$. 
	
	\section{Alternating Optimization} 
	Due to the hidden variables (\textit{i.e.,} unknown true labels), Eq.~\eqref{eq: likelihood function} is difficult to be optimized directly \citep{bishop2006pattern}. Fortunately, based on Jensen's inequality, we can maximize its lower-bound to approach the optimal solution, namely, 
	\begin{equation}
	\underset{\Theta}{\operatorname{argmax}} {\sum_{i\in\left[N\right]} \sum_{j\in\left[c\right]} \mathbf{q}_{i}^{j} \log \hat{P}(\widetilde{\mathbf{y}}_i, \mathbf{y}^j_i=1 | \mathbf{x}_i; \Theta)}, \label{eq: maximizing lower bound}
	\end{equation}
	where $\mathbf{q}_i^j$ is the estimated posterior of the true label defined by $\hat{P}(\mathbf{y}_i^j=1|\widetilde{\mathbf{y}}_i, \mathbf{x}_i;\Theta')$, and $\Theta'$ is the current parameters.
	
	As shown in the following, the optimization target $\mathbf{q}_i$ is related to the learnable parameters in $\Theta$. Therefore, an alternating optimization framework is adopted to solve Eq.~\eqref{eq: maximizing lower bound}: It iteratively estimates the true label posteriors $\mathbf{q}_i$ and adaptively updates the learnable parameters in $\Theta$. The resultant method proceeds to the optimal solution by alternating between two steps, namely, \textit{predicting step} and \textit{updating step}. 
	
	\textbf{Predicting Step} re-estimates the true label posterior $\mathbf{q}_i^j$ by the current parameters. Based on Bayes formula and \mbox{Eq.}~\eqref{eq: assist}, the estimated posterior of true label is:
	\begin{equation}
	\begin{aligned} 
	\mathbf{q}^j_i=
	\frac{1}{K_i}\left[(1-\eta_i)\widetilde{\mathbf{y}}_i^j+\eta_i\psi_i\right]\mathbf{h}_{\mathbf{w}}^j(\mathbf{x}_i)
	, \label{eq 2023}
	\end{aligned}
	\end{equation}
	where $K_i=\sum_{j\in\left[c\right]} P(\widetilde{\mathbf{y}}_i, \mathbf{y}_i^j=1| \mathbf{x}_i;\Theta)$ makes the elements of $\mathbf{q}_i$ sum to 1. Eq.~\eqref{eq 2023} can be rewritten in vector as:
	\begin{equation}
	\mathbf{q}_i=\frac{1}{K_i}\mathbf{h}_{\mathbf{w}}(\mathbf{x}_i)*\left[(1-\eta_i)\widetilde{\mathbf{y}}_i+\eta_i\psi_i\mathbf{1}\right], \label{eq 2213}
	\end{equation}
	where $*$ denotes the element-wise product and $\mathbf{1}$ is a vector of all 1. Therein, to estimate the true label posterior $\mathbf{q}_i$, the noisy label $\widetilde{\mathbf{y}}_i$ is first linearly interpolated with $\mathbf{1}$ according to $\eta_i$ and $\psi_i$, which integrates label ambiguity as in \citep{gao2017deep}. Then, the smoothed label $(1-\eta_i)\widetilde{\mathbf{y}}_i+\eta_i\psi_i\mathbf{1}$ is weighted by $\mathbf{h}_{\mathbf{w}}(\mathbf{x}_i)$ to further combine the classifier prediction, and the normalizer $K_i$ insures that $\mathbf{q}_i\in\mathcal{Q}$ is a valid label distribution.
	
	\textbf{Updating Step} updates trainable parameters $\Theta$ via \mbox{Eq.}~\eqref{eq: maximizing lower bound}, where $\mathbf{q}_i$ is the re-estimated posterior given by Eq.~\eqref{eq 2023}. Ignoring the constant parts, maximizing Eq.~\eqref{eq: maximizing lower bound} w.r.t. the confusing probability $\eta_i$ is equivalent to:
	\begin{equation}
	\underset{0 \leq \eta_i \leq 1,~\forall i}{\operatorname{argmax}} \sum\limits_{i\in\left[N\right]}\sum\limits_{j\in\left[c\right]} {\mathbf{q}_{i}^{j} \log \left[\left(1-\eta_{i}\right)\widetilde{\mathbf{y}}_i^j+\eta_{i}\psi_i\right]}. \label{eq: optimization eta}
	\end{equation}
	Similarly, maximizing  Eq.~\eqref{eq: maximizing lower bound} w.r.t. the DNN parameters $\mathbf{w}$ can be written as:
	\begin{equation}
	\underset{\mathbf{w}}{\operatorname{argmax}} \sum_{i\in\left[N\right]} \sum_{j\in\left[c\right]} \mathbf{q}_i^j \log \mathbf{h}^j_{ \mathbf{w}}\left(\mathbf{x}_i\right), \label{eq: optimization cnn}
	\end{equation}
	which is the canonical maximum likelihood estimation that directly uses true label distributions as optimization targets.
	
	\noindent\textbf{Remarks}. As we can see from Eq.~\eqref{eq 2213}, the noisy label $\widetilde{\mathbf{y}}_i$ is directly taken as the correct one if  $\eta_i=0$. As the value of $\eta_i$ increases, the model prediction $\mathbf{h}_\mathbf{w}(\mathbf{x}_i)$  gradually dominates the true label prediction. Moreover, the model prediction is directly used for estimating the probability of the true label if $\eta_i$ can equal to 1. Generally, $\eta_i$ can be viewed  as a learnable trade-off parameter, automatically determining the degree of learning from the noisy label and the model prediction.
	
	\subsection{Updating Rules}
	
	The constrained optimization Eq.~\eqref{eq: optimization eta} can be solved by Projected Gradient Ascent (PGA) \citep{nocedal2006numerical}, as the constraint set $\left\{\eta_i:0 \leq \eta_i \leq 1\right\}$ is convex in nature. In each iteration, each $\eta_i$ is first updated by gradient ascent:
	\begin{equation}
	\eta_i\leftarrow\eta_i+\alpha_1 \frac{[\mathbf{1}+(\psi_i\eta_i-\eta_i-1)\widetilde{\mathbf{y}}_i]^\top\mathbf{q}_i}{\eta_i + \epsilon} 
	, \label{eq: update eta}
	\end{equation}
	where $\alpha_1$ is the learning rate and  $\epsilon$ is ﬁxed to 0.0001 to avoid dividing by zero. Then, PGA chooses the value of $\eta_i$ that is nearest to the constraint $\{\eta : 0 \leq \eta \leq 1\}$ via:
	\begin{equation}
	\eta_i\leftarrow\min(\max(\eta_i,0), 1), 
	\end{equation}
	such that the updated value of $\eta_i$ in Eq.~\eqref{eq: update eta} still meets the constraint in Eq.~\eqref{eq: optimization eta}. 
	For the DNN parameters $\mathbf{w}$, if gradient ascent is adopted, the updating rule in solving Eq.~\eqref{eq: optimization cnn} is: 
	\begin{equation}
	\mathbf{w}\leftarrow\mathbf{w}+\alpha_2\sum_{i\in \left[N\right]} \sum_{j\in\left[c\right]} \frac{\mathbf{q}_{i}^{j}}{\mathbf{h}^{j}_{\mathbf{w}}\left(\mathbf{x}_{i} \right)} \nabla_{\mathbf{w}} \mathbf{h}^{j}_{ \mathbf{w}}\left(\mathbf{x}_{i}\right), \label{eq: update classifier}
	\end{equation}
	where $\alpha_2$ is the learning rate.
	
	\subsection{Overall Algorithm} 
	
	In alternating optimization, to prevent inaccurate true label estimation at beginning, each confusing probability $\eta_i$ is initialized by a small positive value (\textit{e.g.}, $\eta^{\textrm{INIT}}=0.01$). 
	Accordingly, the estimated label distribution $\mathbf{q}_i$ is dominantly determined by the noisy label to avoid unstable model prediction.
	Besides, it is well-known that DNNs avoid learning label noise in the early training phase~\citep{arpit2017closer,tanaka2018joint}, so this initialization strategy can effectively prevent our classifier from misleading initial training.  
	
	For large-scale learning tasks, such as image classification, the aforementioned alternating optimization is adopted in a mini-batch manner. The resultant learning method is summarized in Algorithm \ref{algorithm}. In each iteration, a mini-batch $\Xi$ is fetched uniformly at random from the training sample. In step 6, we re-estimate true label posteriors by applying Eq.~\eqref{eq 2213} for instances in $\Xi$. In step 8-11, PGA executes (if necessary) one iteration to the subset thereof confusing probabilities only to the mini-batch $\Xi$. In step 12, the DNN parameters $\mathbf{w}$ are updated by mini-batch gradient ascent.

	\begin{algorithm}[t]   
		\caption{The Overall Algorithm.}  
		\label{algorithm} 
		
		\begin{algorithmic}[1]  
			
			\Require the noisy-labeled training sample $(S, \widetilde{Y})$; the estimated probabilities $\{\psi_1,\ldots,\psi_N\}$; number of training steps \textit{num\_steps}; and hyperparameters $\eta^{\textrm{INIT}}$, $\alpha_1$, $\alpha_2$.
			\Ensure the optimal DNN parameters $\mathbf{w}$.
			\State Initialize ${\eta}_i={\eta}^{\textrm{INIT}},\forall i$;
			\State Initialize parameters $\mathbf{w}$; 
			\For{$t\leftarrow1$ to \textit{num\_steps}} 
			\State Fetch a mini-batch $\Xi$ uniformly at random;
			\State \textit{\# predicting step}
			
			\State $\begin{aligned} 
			\mathbf{q}_i = \frac{1}{{K}_i}\mathbf{h}_{\mathbf{w}}\left(\mathbf{x}_{i}\right) *\left[\left(1-\eta_{i}\right) \widetilde{\mathbf{y}}_{i} + \eta_{i}\psi_i \mathbf{1}\right], i \in \Xi
			\end{aligned}$; 
			
			\State \textit{\# updating step}
			
			\If{update posterior}
			\State $\begin{aligned} \eta_i\leftarrow\eta_i+\alpha_1 \frac{[\mathbf{1}+(\psi_i\eta_i-\eta_i-1)\widetilde{\mathbf{y}}_i]^\top\mathbf{q}_i}{\eta_i+\epsilon};
			\end{aligned}$
			\State $\begin{aligned} \eta_i\leftarrow\min(\max(\eta_i,0), 1),i\in\Xi\end{aligned}$;
			\EndIf
			\State $\begin{aligned}
			\mathbf{w} \leftarrow \mathbf{w}+\alpha_{2} \sum_{i \in \Xi} \sum_{j\in\left[c\right]} \frac{q_{i}^{j}}{h^{j}_{\mathbf{w}}\left(\mathbf{x}_{i}\right)} \nabla_{\mathbf{w}} h^{j}_{\mathbf{w}}\left(\mathbf{x}_{i}\right)
			\end{aligned}$; 
			\EndFor				
			\State\Return $\mathbf{w}$
		\end{algorithmic}  
	\end{algorithm}

\section{Experiments}
	
	We first test the proposed method on \textit{CIFAR-100} and \textit{CIFAR-10}~\citep{krizhevsky2009learning} with synthetic label noise, and then conduct real-world label noise experiments on \textit{Clothing1M}~\citep{xiao2015learning}.
	
	\subsection{Datasets}
	
	\noindent\textit{\textbf{CIFAR-100}}: 
	To generate the simulated dataset with instance-dependent label noise, we train a Multi-Layer Preceptron (MLP) on training sample with original labels. Then, we re-label the clean sample by the predicted labels given by MLP. The training accuracy of the MLP is 69.29\%, resulting in a training dataset with label noise. These noisy labels are instance-dependent, as they are determined by the decision boundaries of the MLP in the instance space.
	
	\noindent\textit{\textbf{CIFAR-10}}: We also leverage the classification results with low capacity models to synthesize noisy labels. To exploit a different IDN setting, we apply an unsupervised clustering algorithm to generate noisy labels. First, we map training instances into embedding features that are outputs of pool-5 layer of ResNet-50 \citep{he2016deep} pre-trained on \textit{ImageNet}. Then, we apply k-means++ \citep{arthur2007k} to these embedding features. Finally, instances in the same cluster are uniformly re-labeled by majority voting of their original labels, and the accuracy of the resulting pseudo labels is only 65.75\%. 
	
	Additionally, we conduct synthetic CCN experiments to further demonstrate the universality of our method: noisy labels are generated by mapping TRUCK $\rightarrow$ AUTOMOBILE, BIRD $\rightarrow$ AIRPLANE, DEER $\rightarrow$ HORSE, and CAT $\leftrightarrow$ DOG with different noise transition probabilities $r$. For example, an instance of BIRD has probability $r$ to be mislabeled by AIRPLANE. It mimics the confusion of similar classes, and the label noise is class-conditional here. 
	
	\noindent\textbf{\textit{Clothing1M}}: \textit{Clothing1M} is a large-scale dataset with noisy \mbox{labels}. It contains more than one million images of clothing crawled from several shopping websites, which are classified into 14 classes. \mbox{Labels} are automatically generated according to surrounding texts of these instances. It is reported that the labels suffer from instance-dependent noise \citep{xiao2015learning}. Hence, we conduct experiments on \textit{Clothing1M} to examine the performance of our method in a real-world IDN setting. The overall label precision for the training dataset is approximately 60\%, and it also contains 50k, 14k, and 10k correctly labeled instances for auxiliary training, validation, and test. Besides, this dataset exhibits serious data imbalance phenomenon. For example, more than 88k instances are labeled to SHIRT, but, in contrast, only 19k instances have the label SWEATER.
	
	\subsection{Implementation Details~\footnote{Note that, same backbone models and hyperparameters are used for directly training DNNs with noisy labels in estimating the probability of the observed noisy labels, \textit{i.e.,} $\{\psi_1,\ldots,\psi_N\}$.}} \label{implementation}

	\textbf{\textit{CIFAR-100} \& \textit{CIFAR-10}}: We employ ResNet-32~\citep{he2016deep} for fair comparison with existing methods. Mean-subtraction, horizontal random flip, and $32 \times 32$ random crop are performed for data pre-processing. Then, we use mini-batch gradient ascent with a momentum of 0.9; a weight decay of $10^{-4}$; and a batch size of 256. The alternating optimization algorithm is executed for 160 epochs. For the classifier parameters, we begin with a learning rate $\alpha_2=0.05$  and divide it by 10 per 40 epochs. 
	For label confusing probabilities, they are updated every 5 epochs from the $35^\text{th}$ epoch. We test model performance with different learning rate $\alpha_1$ on \textit{CIFAR-100}, and fix $\alpha_1=0.7$ on \textit{CIFAR-10}.
	
	\noindent\textbf{\textit{Clothing1M}}: As a common setting, ResNet-50 pre-trained on \textit{ImageNet} is utilized as the backbone model. In addition to the common data pre-processing techniques (\textit{i.e.}, mean subtraction, horizontal random flip, and $224 \times 224$ random crop), oversampling is applied by adding more copies for training instances that are labeled to each minority class. Oversampling is widely used for tackling data imbalance problem, while the oversampled dataset is not truly balanced, because the observed labels are not the true labels. We use mini-batch gradient ascent with a momentum of 0.9, a weight decay of $10^{-3}$, and a batch size of 32. The alternating optimization algorithm is executed for 15 epochs. For the label confusing probabilities, their values are updated from the second epoch with a fixed learning rate $\alpha_1=0.05$. For the classifier parameters, the learning rate $\alpha_2$ is set to be $5\times10^{-3}$ and divided by 10 per 5 epochs.
	
	\subsection{The Adopted Baseline Methods}
		
	We compare with the following label noise learning algorithms in our experiments: ``Co-teaching'' \citep{han2018co} is the state-of-the-art sample selection algorithm which can avoid distribution bias; ``Forward'' \citep{patrini2017making} and ``$\mathcal{L}_\text{DMI}$'' \citep{xu2019l_dmi} are two noise-robust methods under CCN assumption;	``Bootstrapping'' \citep{reed2014training} is the pioneer that handles the label noise by label correction; and ``Tanaka'' \citep{tanaka2018joint} as well as ``PENCIL'' \citep{yi2019probabilistic} are two advanced methods that also use model predictions in correcting noisy labels. 
	
	There have been recent advances that incorporate the methodology of meta-learning~\citep{ren2018learning,shu2019meta} and designing specific network architectures~\citep{lee2018cleannet}. However, for fair comparison, their results are not listed here as they typically rely on additional training instances with manually verified labels. 
	Besides, for fair comparison, we re-implement ``Bootstrapping'' and ``Co-teaching'' by using ResNet-32 on \textit{CIFAR-100} and \textit{CIFAR-10}; and using ResNet-50 on \textit{Clothing1M}.
	
	\subsection{Experiments on \textit{CIFAR-100}}

	\begin{table}[t]
	    \centering

	    \small
	    \begin{tabular}{c||c|c}
	        \toprule
	         & Method & Test Accuracy (\%) \\
	         \hline\hline
	         \#1 & \tabincell{c}{Co-teaching\\ \citep{han2018co}} & 45.15$\pm$0.53\\ 
	         \hline
	         \#2 & \tabincell{c}{Forward\\ \citep{patrini2017making}} & 44.97$\pm$0.77\\
	         \hline
	         \#3 & \tabincell{c}{$\mathcal{L}_\text{DMI}$\\ \citep{xu2019l_dmi}} & 45.07$\pm$0.42\\
	         \hline
	         \#4 & \tabincell{c}{Bootstrapping\\ \citep{reed2014training}} & 44.52$\pm$0.35 \\
	         \hline
	         \#5 & \tabincell{c}{Tanaka\\ \citep{tanaka2018joint}} & 46.02$\pm$0.42\\
	         \hline
	         \#6 & \tabincell{c}{PENCIL\\ \citep{yi2019probabilistic}} & 45.57$\pm$0.40 \\
	         \hline
	         \#7 & Ours & \textbf{47.51$\pm$0.28} \\
	        \toprule
	    \end{tabular}
	    \caption{Average test accuracy and standard deviation (5 trials) on \textit{CIFAR-100} with synthetic label noise.}
	    \label{table:cifar100}
	\end{table}
	
	\begin{table*}[t]		
		\centering
		
		\label{table:cifar10}
	    \small
	    \begin{tabular}{c||c|c|ccccc} 
		\toprule
			\multirow{2}*{} & \multirow{2}*{Method} & \multirow{2}*{{IDN}(\%)} &  \multicolumn{5}{|c}{{CCN}(\%)} \\
			\cline{4-8}
			& & & 0.1 & 0.2 & 0.3 & 0.4 & 0.5\\
			\hline\hline
			\#1 & Co-teaching &  {66.93$\pm$0.21}  & {91.31$\pm$0.05} & {89.01$\pm$0.14} & {83.56$\pm$0.30} & {79.67$\pm$0.59} & {73.30$\pm$0.41}\\
			\hline
			\#2 & Forward &  {64.29$\pm$1.07}  & {91.55$\pm$0.08} & {90.64$\pm$0.12} & {87.46$\pm$0.17} & {85.65$\pm$0.23} & \textbf{{80.22$\pm$0.39}} \\
			\hline
			\#3 & $\mathcal{L}_\text{DMI}$ & 66.15$\pm$0.43 & 91.73$\pm$0.05 & 90.07$\pm$0.11 & 91.25$\pm$0.35 & 88.80$\pm$0.27 & 76.15$\pm$0.37 \\
			\hline
			\#3 & Bootstrapping &  {66.24$\pm$0.44}  &  {88.53$\pm$0.13} & {84.01$\pm$0.13} & {80.53$\pm$0.20} & {78.95$\pm$0.17} & {72.07$\pm$0.83}  \\
			\hline
			\#4 & Tanaka & {67.77$\pm$0.57} & {92.19$\pm$0.02} & {91.97$\pm$0.13} & {91.64$\pm$0.11} & {90.39$\pm$0.16} & {68.39$\pm$1.25}\\
			\hline
			\#5 & PENCIL &  {69.51$\pm$0.63} & {93.27$\pm$0.10} & {92.86$\pm$0.15} & {91.29$\pm$0.20} & {89.25$\pm$0.23} & {76.18$\pm$0.73}\\
			\hline
			\#6 & {Ours} &  \textbf{{69.82$\pm$0.20}}  & \textbf{{93.81$\pm$0.05}}  & \textbf{{93.06$\pm$0.10}} & \textbf{{92.79$\pm$0.21}} & \textbf{{90.48$\pm$0.16}} & {78.03$\pm$0.55}\\
        \toprule
		\end{tabular}
		\caption{Average test accuracy and standard deviation (5 trials) on \textit{CIFAR-10} with synthetic label noise.}
	\end{table*}
	
	\begin{figure}[t]
	    \centering
	    \includegraphics[width=2.3in]{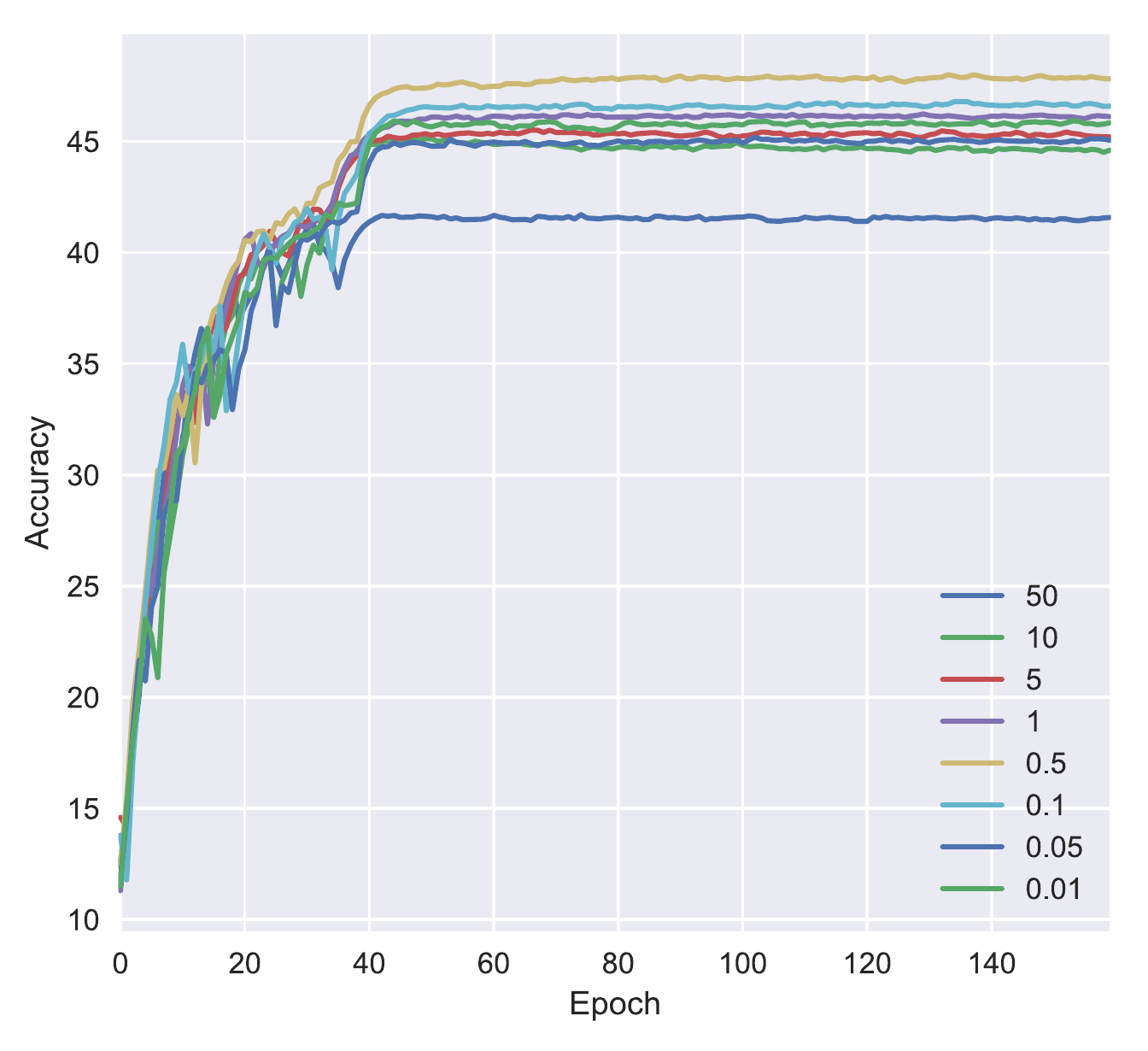}
	    \caption{Average validation accuracy curves (5 trials) on \textit{CIFAR-100} with different learning rate $\alpha_2$ of the confusing probabilities. Note that, the standard deviations are not provided for clarity.}
	    \label{fig:cifar100_curve}
	\end{figure}
	
	To evaluate the performance of our method in IDN, we first test our proposed method on \textit{CIFAR-100} with synthetic label noise. Our method and the baselines are implemented five times, and the average results are reported in Table~\ref{table:cifar100}. As we can see, ``$\mathcal{L}_\text{DMI}$'' and ``Forward'' reveal relatively inferior performance due to their non-adaptiveness to IDN cases.
    The performance of ``Bootstrapping'' is also not effective, in part because of its inherent difficulty in hyperparameter tuning. 
    By contrast, ``Co-teaching'' and two model prediction based methods ``Tanaka'' and ``PENCIL'' show much preferred results, while our method still outperform them by 1.49\% to 1.94\%. These results clearly demonstrate the merits of specifying the noisy label generation process in handling IDN. 
	
	In our method, the learning rate $\alpha_2$ for confusing probabilities plays a key role in tackling the label noise. To provide some guidelines for hyperparameter tuning, we show the validation accuracy curves during training in Fig.~\ref{fig:cifar100_curve}, given the learning rates $\alpha_2$ with different orders of magnitude. At the beginning, all the curves precariously vibrate due to the relatively large initial learning rate, while they are suddenly increase and become stable when the confusing probabilities start to update near $35^\text{th}$ epoch. As we can see, the performance reaches the highest result when the learning rate equals to 0.5. Moreover, with the rapidly increase of $\alpha_2$, the model performance drops quickly (\textit{e.g.}, $\alpha_2 = 50$). The reason is that, with a large learning rate, too many instances are taken as confusing at beginning, and the imperfect model predictions might be directly taken as the true label distributions. By contrast, though validation accuracy with extremely small learning rates (\textit{e.g.}, $\alpha_2 = 0.01$) can also hurt the model performance, the impact is not as severe as the large cases. Here, most of the instances are taken as unconfusing ones, and it simply degenerates to a naive situation that learns directly from the noisy labels.

	\subsection{Experiments on \textit{CIFAR-10}}
	
	We further conduct IDN experiments on \textit{CIFAR-10}. The average results of five individual trials are summarized in the third column of Table~\ref{table:cifar10}. According to the results, our method achieves overall higher test accuracy, again demonstrating the superiority of our learning method in handling IDN situations. The advanced algorithms implicitly handle instance-dependent noise based on heuristic rules, \textit{e.g.}, self-learning and small-loss assumption, while our method still outperforms these baselines by 0.31\% to 3.58\%. By contrast, two CCN based methods, \textit{i.e.}, ``$\mathcal{L}_\text{DMI}$'' and ``Forward'', perform relatively inferior to other baselines and our method, which reflects the significance in studying IDN cases. 
	
	Additionally, to show the proposed method can also properly handle class-conditional label noise, we conduct CCN experiments with noise transition probabilities $r$ ranging from 0.1 to 0.5. Table~\ref{table:cifar10} reports the average empirical results from column 3 to 7. Our setup is generally in line with \citep{yi2019probabilistic}. However, to demonstrate the robustness of each method, we do not adaptively change their hyperparameters under various $r$ cases. Compared with ``Forward'' and ``$\mathcal{L}_\text{DMI}$'', two robust learning algorithms in handling CCN, our method can outperform their results in most cases. The reason for the superior performance of ``Forward'' when $r=0.5$ is that it uses the ground-truth noise transition matrix, which is unrealistic in real situations. Our method outperforms ``Co-teaching'' and ``Bootstrapping'', of which the results show extreme overfitting phenomenon. Furthermore, the average test accuracy of our method is also slightly better than ``Tanaka''  as well as ``PENCIL'', which are two state-of-the-art methods in these settings.  
	
	\subsection{Experiments on \textit{Clothing1M}}
    To evaluate the  proposed method in a real-world setting, we conduct experiments on \textit{Clothing1M}, and the results are summarized in Table~\ref{table:clothing1m}. Row \#2 and \#5 are respectively quoted from \citep{patrini2017making} and \citep{tanaka2018joint}, and their results both rely on side information. In Row \#2, Patrini {et al.} exploit 50k extra clean training instances to estimate the noise transition matrix. In Row \#5, Tanaka {et al.} use the distribution prior of true labels to relieve the class imbalance problem. In contrast, we do not rely on any side information, while our method (Row \#7) achieves 1.79\% better test accuracy than that of ``Tanaka'' and 4.18\% higher than ``Forward''. Our method also achieves substantial improvement over ``Bootstrapping'', ``Co-teaching'', and ``$\mathcal{L}_\text{DMI}$'', and pushes the best test accuracy reported by ``PENCIL'' from 73.49\% to 74.02\%. 
	
	We further exploit 50k training instances with correct labels by mixing them with the noisy labelled training sample. To make fully use of these clean labels, each mini-batch consists of the same number of instances drawn from the noisy-labeled and clean training instances. The confusing probabilities for the clean instances are fixed to 0 throughout the training procedure, and the result is reported in Row \#8. As we can see, there is a noticeable improvement compared with the result in Row \#7 that only utilizes the noisy-labeled instances. Obviously, training instances with correct labels can provide much reliable guide information during the training procedure. Besides, the clean training instances can also be used for finetuning the resulting classifier in \#7, and the new state-of-the-art \#10 outperforms the similar finetuned result of ``Forward'' in \#9 more than 0.3\%. 

\begin{figure}[t]
	    \centering  
		\subfigure[1M]{
		    \centering  
			\begin{minipage}[t]{0.45\linewidth}
			    \centering
				\includegraphics[width=2in]{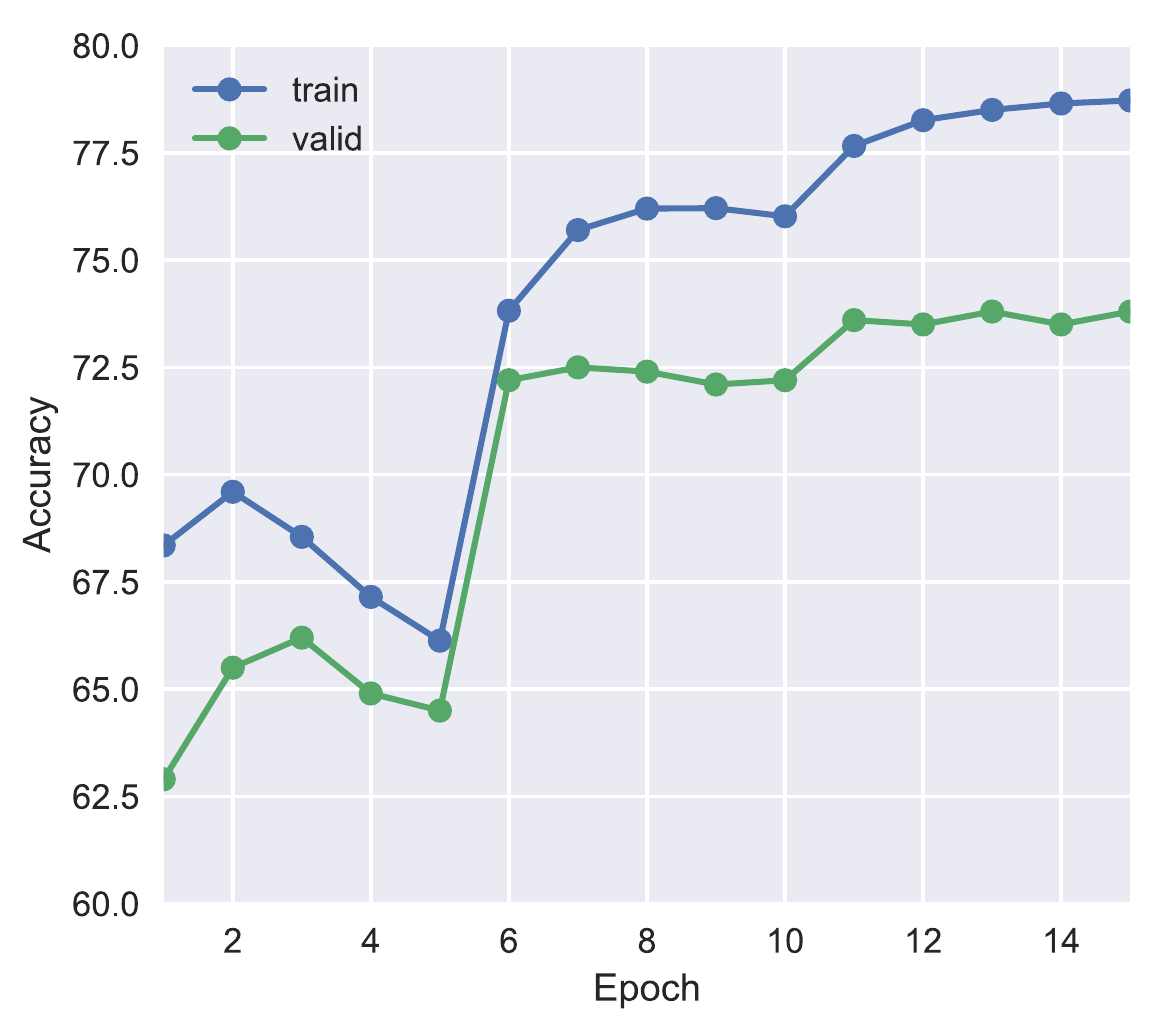}
				\centering  
				\label{curve_noise}
		\end{minipage}} 
		\subfigure[1M + 50k]{
		    \centering  
			\begin{minipage}[t]{0.45\linewidth}
			    \centering  
				\includegraphics[width=2in]{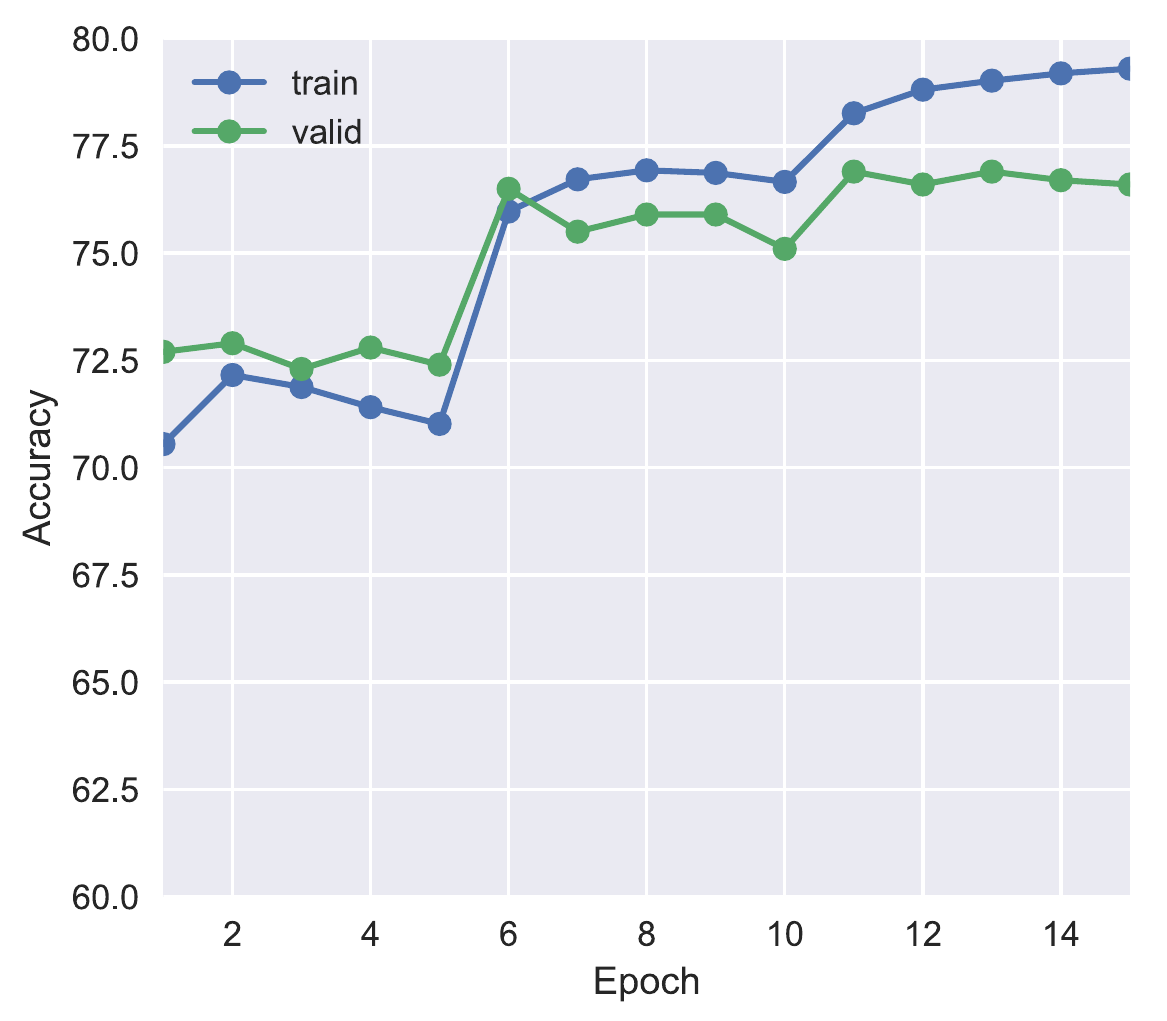}
				\centering  
				\label{curve_clean}
		\end{minipage}}
		\caption{Training and validation accuracy curves on \emph{Clothing1M} with/without clean data.} 
		\label{curve}
	\end{figure}

	To provide some insights into the learning behavior of the classifier in our method, we plot validation and training accuracy curves in Fig.~\ref{curve} with two different learning strategies. Note that the training accuracy is calculated w.r.t. the noisy labels, instead of the clean labels as in validation. Abnormally high training accuracy typically indicates that the classifier overfits the noisy labels, while we always wish the validation accuracy is as high as possible. Fig.~\ref{curve_noise} reports the result that only uses noisy-labeled instances. The training and validation accuracy both increase in the first two epochs, since the noisy labels are directly taken as the correct ones for true label posteriors. Then, model predictions gradually participate into the prediction of true label posteriors. Accordingly, the training accuracy curve declines sharply, while the validation accuracy curve keeps more steady. This indicates that our method can effectively avoid the classifier in overfitting  the label noise. Subsequently, both curves increase with the decrease of the learning rate, and finally become stable. Similar shape of the accuracy curves can be observed in Fig.~\ref{curve_clean}, of which the result also exploits 50k clean training instances as in \#8. Although the training accuracy in both cases are stable at nearly 80\%, there is an obvious improvement for the validation accuracy curve in Fig.~\ref{curve_clean}. This demonstrates the benefit of utilizing clean labels in training classifiers. 
	
	\begin{table}[t]
		
		\centering  

		\label{table:clothing1m}  
        \small
		\begin{tabular}{c||c|c|c}  
	        \toprule
			& Method & \tabincell{c}{Test\\Accuracy(\%)} & \tabincell{c}{Side\\Information} \\
			\hline\hline
			\#1 & Co-teaching & 66.35 &  - \\
			\#2 & Forward & 69.84 &  transition matrix \\
			\#3 & $\mathcal{L}_\text{DMI}$ & 72.46 & - \\
			\#4 & Bootstrapping & 67.55 & - \\
			\#5 & Tanaka & 72.23 &  label distribution \\
			\#6 & PENCIL & 73.49 & -\\
			\#7 & Ours & \textbf{74.02} & - \\
			\hline
			\#8 & Ours & \textbf{77.55} & +50k (train) \\
			\#9 & Forward & 80.38 &  +50k (finetune) \\
			\#10 & Ours & \textbf{80.68} & +50k (finetune) \\
			\toprule
		\end{tabular}  
		\caption{Similar to previous works, we report the best test accuracy on \textit{Clothing1M}. \#2, \#9 are quoted from \citep{patrini2017making}, \#5 is quoted from \citep{tanaka2018joint}, and \#6 is quoted from \citep{yi2019probabilistic}. }
	\end{table}  

	\begin{figure}[t]
		\centering
		\includegraphics[width=2.7in]{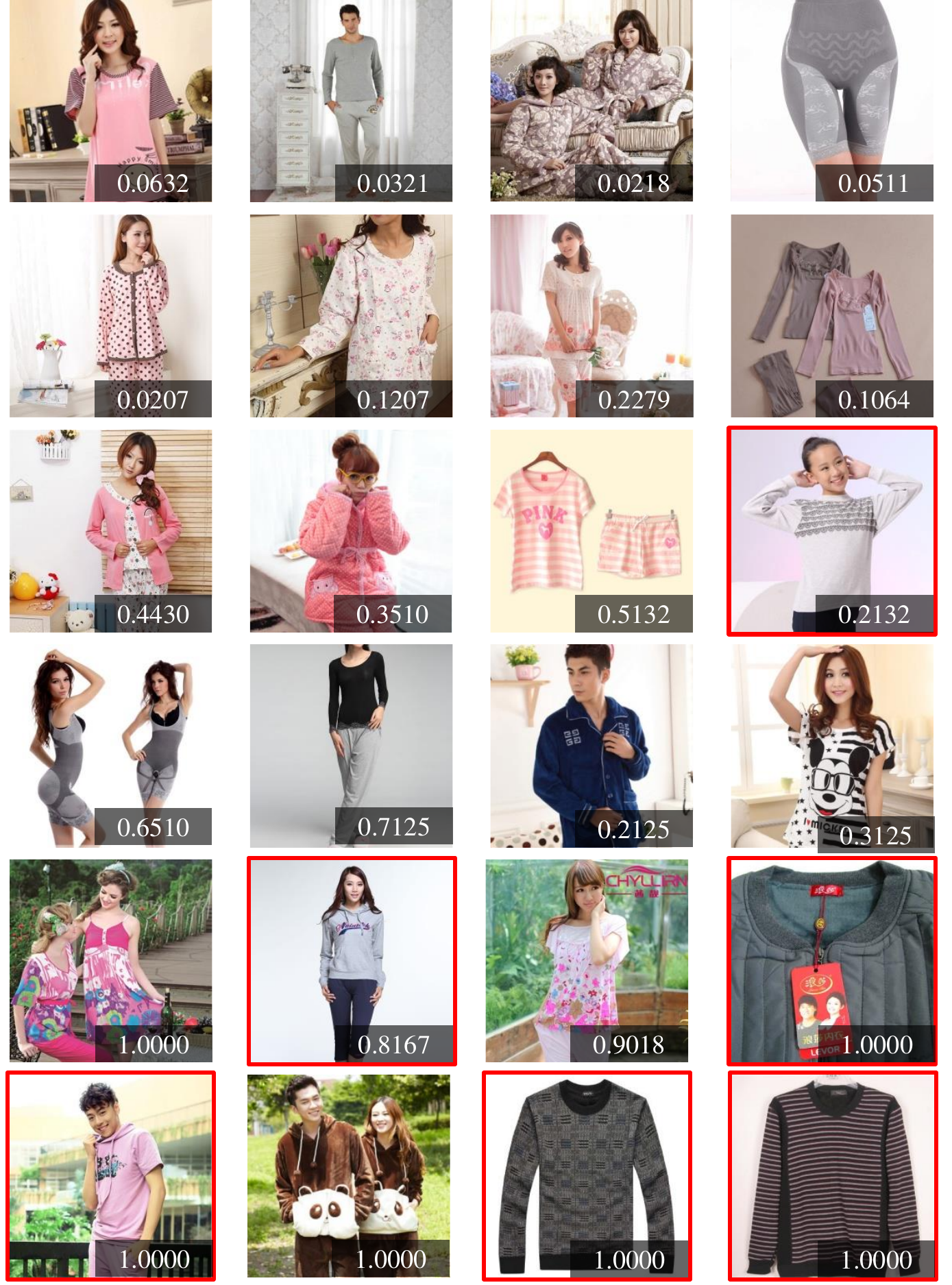}
		\caption{Examples of the training instances with label UNDERWEAR on \textit{Clothing1M}. The estimated confusing probabilities are in the bottom right corner, and the mislabeled instances are in red box.}
		\label{fig:cloth} 
	\end{figure}
	
	In Fig.~\ref{fig:cloth}, we give examples of the training instances whose noisy labels are assigned to UNDERWEAR with the estimated confusing probabilities in their lower right corner. For instances with relatively small confusing probabilities in the first two rows, they are all correctly labeled, and can roughly be viewed as unconfusing instances. With larger confusing probabilities in the third and fourth rows, the corresponding instance pattern becomes diverse, which also contain an instance that is mislabeled. Finally, for instances with extremely large confusing probabilities in the last two rows, instances with rare patterns and incorrect labels become \mbox{common}.

	\section{Conclusion}
	By categorizing training instances into confusing and unconfusing instances, we propose a novel probabilistic model to describe the generation of instance-dependent label noise. Our model is realized by DNNs with additional trainable parameters in finding the potentially confusing instances, and we deploy an alternating optimization algorithm to iteratively correct noisy labels and update trainable parameters. The learning behavior is explainable, which can correct the noisy labels of confusing instances according to the model predictions. In the future, we will focus on the estimation algorithms for confusing probabilities, and extend our model to more weakly supervised learning scenarios \citep{kiryo2017positive,yu2018learning}.
	
	\section{Acknowledgments}
	BH was supported by the RGC Early Career Scheme No. 22200720, NSFC Young Scientists Fund No. 62006202, HKBU Tier-1 Start-up Grant, HKBU CSD Start-up Grant, and HKBU CSD Departmental Incentive Grant. TLL was supported by Australian Research Council Project DE-190101473. JY was supported by NSFC No. U1713208 and “111 Program”under Grant No. AH92005. CG was supported by NSFC No. 61973162, the Fundamental Research Funds for the Central Universities No. 30920032202, CCF-Tencent Open Fund No. RAGR20200101, the “Young Elite Scientists Sponsorship Program” by CAST No. 2018QNRC001, and Hong Kong Scholars Program No. XJ2019036.

	\bibliography{reference}

\end{document}